\documentclass[twocolumn, switch]{article}

\usepackage{arxiv}

\usepackage[utf8]{inputenc} % allow utf-8 input
\usepackage[T1]{fontenc}    % use 8-bit T1 fonts
\usepackage{hyperref}       % hyperlinks
\usepackage{url}            % simple URL typesetting
\usepackage{booktabs}       % professional-quality tables
\usepackage{amsfonts}       % blackboard math symbols
\usepackage{nicefrac}       % compact symbols for 1/2, etc.
\usepackage{microtype}      % microtypography
\usepackage{lipsum}
\usepackage[binary-units=true]{siunitx}
\usepackage{comment}
\usepackage{stfloats}
\usepackage{graphicx}
\usepackage{multirow}
\usepackage{makecell}
\usepackage{epsfig}
\usepackage{pifont}
\usepackage{booktabs} % For formal tables
\usepackage{footnote}
\usepackage[english]{babel}
\usepackage[linesnumbered,algoruled]{algorithm2e}
\usepackage{amsmath}
\usepackage{bm}
\usepackage{multicol}
\usepackage{amssymb}
\usepackage{setspace}
\usepackage{tikz}

\usepackage{xcolor}
\usepackage{cite}
\usepackage{subfigure}
\usepackage{textcomp}
\usepackage{comment}
\usepackage{xcolor}
\usepackage[export]{adjustbox}

\newcommand\copyrighttext{%
  \footnotesize \textbf{IEEE Copyright Notice.}
\textcopyright 2019 IEEE. Personal use of this material is permitted. Permission from IEEE must be obtained for all other uses, in any current or future media, including reprinting/republishing this material for advertising or promotional purposes, creating new collective works, for resale or redistribution to servers or lists, or reuse of any copyrighted component of this work in other works.}
\newcommand\copyrightnotice{%
\begin{tikzpicture}[remember picture,overlay]
\node[anchor=south,yshift=10pt] at (current page.south) {\fbox{\parbox{\dimexpr\textwidth-\fboxsep-\fboxrule\relax}{\copyrighttext}}};
\end{tikzpicture}%
}

\title{TentacleNet: A Pseudo-Ensemble Template for Accurate Binary Convolutional Neural Networks\vspace*{-0mm}}

\date{\vspace{-5ex}}

\author{
Luca Mocerino, Andrea Calimera\\
Politecnico di Torino, 10129 Torino, Italy\vspace*{0mm}
}

\begin{document}
\twocolumn[ % Method A for two-column formatting
  \begin{@twocolumnfalse} % Method A for two-column formatting

\maketitle

%\copyrightnotice
\vspace{-1cm}

\begin{abstract}
 Binarization is an attractive strategy for implementing lightweight Deep Convolutional Neural Networks (CNNs). Despite the unquestionable savings offered, memory footprint above all, it may induce an excessive accuracy loss that prevents a widespread use. This work elaborates on this aspect introducing TentacleNet, a new template designed to improve the predictive performance of binarized CNNs via parallelization. Inspired by the {\em ensemble learning} theory, it consists of a compact topology that is end-to-end trainable and organized to minimize memory utilization. Experimental results collected over three realistic benchmarks show TentacleNet fills the gap left by classical binary models, ensuring substantial memory savings w.r.t. state-of-the-art binary ensemble methods.
\end{abstract}
\vspace{0.1cm}

% keywords can be removed
\vspace{0.35cm}

%\textbf{IEEE Copyright Notice.}
%\textcopyright 2019 IEEE. Personal use of this material is permitted. Permission from IEEE must be obtained for all other uses, in any current or future media, including reprinting/republishing this material for advertising or promotional purposes, creating new collective works, for resale or redistribution to servers or lists, or reuse of any copyrighted component of this work in other works. \\

\textbf{This is a Preprint}. Accepted to be Published in Proceedings of \textcopyright 2020 IEEE International Conference on Artificial Intelligence Circuits and Systems, March 23-25 2020, Genova, Italy.

\copyrightnotice

\vspace{0.35cm}

\section{Introduction}

\end{@twocolumnfalse}]
%Convolutional Neural Networks (CNNs) are supervised inference models that reached human-level accuracy in perception. They are particularly suited for unstructured data, like images, audio and text, therefore finding application in the fields of computer vision, speech recognition and natural language processing~\cite{cnn_general}. 
Convolutional Neural Networks (CNNs) are known to be highly redundant, a positive characteristic for the training because it helps to achieve higher accuracy, but highly undesired during inference, when extra-functional metrics, like latency, energy and memory footprint, are just as important. No matter if the target is a cloud application hosted on a server queried by millions of users, or a mobile application run on low-power cores with limited resources, an efficient use of CNNs calls for effective optimization strategies.% that can reduce the resource usage.

There is plenty of compression or approximation techniques that serve this purpose operating at different levels of abstraction which leverage different knobs~\cite{sze2017efficient,han2016eie,mocerino2019energy}. At the bit-level, {\em binarization} is a very attractive option. The pioneering idea, firstly introduced in~\cite{BinCon} and then elaborated in~\cite{BNN} and~\cite{xnor}, is to project weights and/or activations into a binary space. Moving from multi-bit representations (either floating-point or fixed-point) to single-bit has clear advantages, such as the lowering of the memory footprint and a better use of the available bandwidth since operands can be packed in a single line and accessed in parallel. Moreover, it allows the replacement of real and integer arithmetic with bit-wise operators, e.g. parallel Boolean XNOR and pop-counting~\cite{xnor}, which are faster and less resource demanding. This latter aspect is however influenced by the type of hardware available. 
%cannot be taken for granted since strongly influenced by the available hardware.
General purpose architectures grew to support mainstream applications operating on single-precision floating-point, therefore, less frequent instructions and unusual data representations, like those deployed in binary CNNs, have been dropped for the sake of area efficiency. To fill this gap, software macros can be used to unpack data and properly feed the execution units. This may introduce substantial performance overhead~\cite{bitflow}. Dedicated hardware accelerators, like those introduced in~\cite{finn,xnorbin,li20177,andri2016yodann}, may be a better option as they can push binary CNNs toward impressive speed-up.
% (up to $58\times$).% w.r.t. full-precision).

In spite of the potential savings brought, the use of binary CNNs is still quite limited, sometimes prohibitive, because of the poor predictive quality. For instance, compared to full-precision (32-bit floating-point), the accuracy drop may range from 2\% to 10\%, but even more depending on the complexity of the task~\cite{xnor}. The objective of this work is to address this limitation introducing a new model template named {\em TentacleNet}. 
The basic working principle is inspired by the {\em ensemble learning} theory, well known in machine and deep learning~\cite{ensemble}, that is, the assembly of many {\em weak} classifiers enables a {\em strong} predictor with higher accuracy. However, TentacleNet shows distinctive features that have been specifically designed to leverage the power of binary BNNs and to optimize resource usage. % for the deployment of binary CNNs as weak classifiers.
Moreover, it is end-to-end trainable and can be applied to any generic CNN model using the training procedures available in common deep learning frameworks. Due to its parallel topology, TentacleNet is ready for the forthcoming generation of parallel architectures with heterogeneous accelerators~\cite{garofalo2019pulp}.

Experimental results collected over three computer vision tasks, i.e. image classification on CIFAR-10 and CIFAR-100~\cite{cifar} and facial expression recognition on the FER13 data-set~\cite{fer13}, reveal TentacleNet can reach the accuracy of full-precision models, yet ensuring much lower memory footprint compared to state-of-the-art binary ensemble methods~\cite{binen}.

\section{Background and Previous Works}\label{sec:back}

\subsection{Binarized Neural Networks}
The recent literature shows several binarization methods for CNNs. In~\cite{BinCon}, Courbariaux et al. introduced the concept of CNNs with binary weights in the range \{-1, 1\}, leaving the activations in full-precision (floating-point 32-bit). In~\cite{BNN}, the same authors presented a full-binary CNN where also the activations get projected in a binary space using a {\em sign} activation function. % instead of the {\em hard sigmoid}.
That is the first example of a full-binary CNNs processed with bit-wise XNOR and pop-counting, with no floating-point arithmetic. As side effect, the prediction accuracy suffered substantial degradation $(\ge 10\%)$. Later, M. Rastegari et. al. presented XNOR-Net~\cite{xnor}, an alternative architecture to mitigate the accuracy drop by re-scaling the binary output of each convolutional layer through a full-precision normalization layer. The improvement over~\cite{BNN} was remarkable: up to 16.3\% on ImageNet~\cite{imagenet}. The XNOR-Net represents still today the state-of-the-art for binary CNNs and therefore we borrowed the same architecture in this work (simply referred as BNN hereafter). However, our strategy can be extended to any type of binarized network.

\subsection{Feature Extraction in a BNN}
Given $\textbf{x} \in \mathbb{R}^{ch \times w_{in}\times h_{in}}$ as the input feature and $\textbf{w} \in \mathbb{R}^{ch \times kw\times kh} $ as the weight tensor, their convolution is approximated as follows:
\begin{equation}
 \textbf{x} \ast \textbf{w} \approx \text{popcount}\;(\textbf{X}\;  \text{xnor}\;  \textbf{W}) \cdot K \cdot \alpha
\end{equation}
where $K$ and $\alpha$ are the scaling factors. While weights (\textbf{W}) are binarized off-line, the binary activation function is fused with the batch normalization and hence run on-line. Given the parameters of the batch normalization, i.e. variance $\sigma^2$, mean $\mu$, scale $\gamma$, shift factor $\beta$, $\epsilon$ a coefficient for numerical stability, a generic feature map {\bf x} is binarized as follows: 
\begin{equation}
    \text{\bf X} = BinACT_{0,1} (x) = \begin{cases}
                \text{1 }  x \ge c \\
                 \text{0 } x < c \\
                \end{cases}
\end{equation}
with $c=\mu-\beta/\gamma \sqrt{\sigma^2 + \epsilon}$. As additional details, we do not use $K$, that is the activations scaling factor, in order to speed-up the training stage; $c$ and $\alpha$ are represented as 32-bit floating point numbers.

It is worth to notice that within a BNN, the first and the last layers are kept and processed to full-precision (floating-point 32-bit) in order to mitigate the accuracy drop induced by the binarization of the inner layers. This is a relevant characteristic exploited by TentacleNet.

The efficient processing of a BNN requires an effective implementation of parallel bit-wise XNOR, pop-counting, and bit-2-word packing/unpacking (e.g. from 1 to 32-bit and vice-versa). While new specialized cores have an extended instruction-set coupled with dedicated hardware units, e.g.~\cite{finn}, for many general-purpose cores the only viable option is to make custom software macros. The performance gap between hardware acceleration and software implementation is large, with the latter being much slower~\cite{bitflow}. For instance, in~\cite{cpu_fpga} Moss et al. showed that a custom FPGA-based inference engine gets 8.5$\times$ faster and 20$\times$ more energy efficient. TentacleNet is orthogonal to the kind of hardware, but it would benefit most from custom accelerators.

\subsection{Ensembles Learning}
Ensemble methods are well-known tools in statistics and machine learning; they are commonly used to improve resilience against under-/over-fitting~\cite{ensemble}. The basic principle is simple, yet effective: use multiple {\em weak} estimators to build up with a single {\em strong} classifier. Random forests are practical examples, where the weak classifiers consist of decision trees~\cite{random_forest}.

Existing ensemble strategies mainly differ on $(i)$ the training procedure adopted and $(ii)$ how the outcome of the weak estimators are grouped and evaluated. The taxonomy is as follows:\\
\textbf{Bagging}~\cite{bagging}. The training dataset $D$ is randomly partitioned into $N$ sub-sets $d_i$ ($i \in [1,N]$), with $N$ the number of weak estimators. Each weak estimator is trained using $d_i$ as the training set. During inference, the outputs of the $N$ estimators are averaged or evaluated with a voting mechanism.\\
%This approach mainly reduces the variance.\\
\textbf{Boosting}~\cite{boost1,boost2}. 
% includes a class of a meta-algorithms for training several {\em weak} learners in order to create a single learner~\cite{boost1,boost2}. 
Each weak estimator is trained (separately) over the full dataset $D$. The outputs of the $N$ estimators are then fused using a linear transformation whose coefficients are learned at training time, for instance using AdaBoost algorithm~\cite{adaboost}.\\
%Several variants differ by their choice of loss function (LogitBoost\cite{logitboost}, L2Boost\cite{l2boost} and LPBoost\cite{lpboost}). This approach aims to reduce both bias and variance.   \\
\textbf{Stacking}~\cite{stack1,stack2}. 
The {\em N} weak estimators are trained on the original data $D$. Then, their outputs are used as training-set for an additional meta-estimator, which is run in sequence during inference. The key feature is that the stack is built upon heterogeneous estimators.

There exist different works that proposed the use of ensemble methods for deep neural networks. Remarkable results are reported in~\cite{cnn_boosted}, where the authors adopted a boosting strategy on image classification, but also in CoopNet\cite{coopnet} which combines multiple precision models to improve accuracy and inference latency.
%in~\cite{ferensemble} which reports on practical example of meta-data enhanced stacking.
%In~\cite{ju2018relative}, Ju et al. introduced a cross-validated version of {\em SuperLearner} achieving better results w.r.t. previous approaches.
Even more interesting, the concept of ensemble learning can be found in the internal architecture of the most recent CNN models. For instance ResNet~\cite{resnet}, DenseNet~\cite{densenet} and Inception series~\cite{googlenet} have layers which combine branches produced by the previous layers to improve performance.
This resembles an ensemble learning structure indeed.
 %or that concatenate different branches

All the above methods were thought to improve accuracy, with no particular attention to the complexity of the weak classifiers. The result is a dramatic increase in the model size. When extra-functional metrics (e.g. memory and latency) enter the cost function, the selection of the weak estimators should be resource-oriented and not just accuracy-driven. In this regard, binary CNNs are good candidates: they are weak, fast and small. Some recent works explored this option. For instance, in~\cite{binen} the authors adapted the classical ensemble methods to binary CNNs, {\em bagging} and {\em boosting} in particular. The collected results revealed that a large number of BNNs is needed to get close to the accuracy of the full-precision model, thus resulting in large memory space. TentacleNet takes a step forward, showing that binary ensembles can reach high accuracy with fewer resources.
    
\section{TentacleNet}
\subsection{Architecture}\label{sec:arch}
TentacleNet is a parallel template embedding lightweights BNNs in a pseudo-ensemble structure. It serves any kind of feed-forward CNN, namely, any full-precision CNN can be translated following the TentacleNet template and exploit the binary computation. 

\begin{figure}[!tt]
\centering
    \includegraphics[scale=0.45]{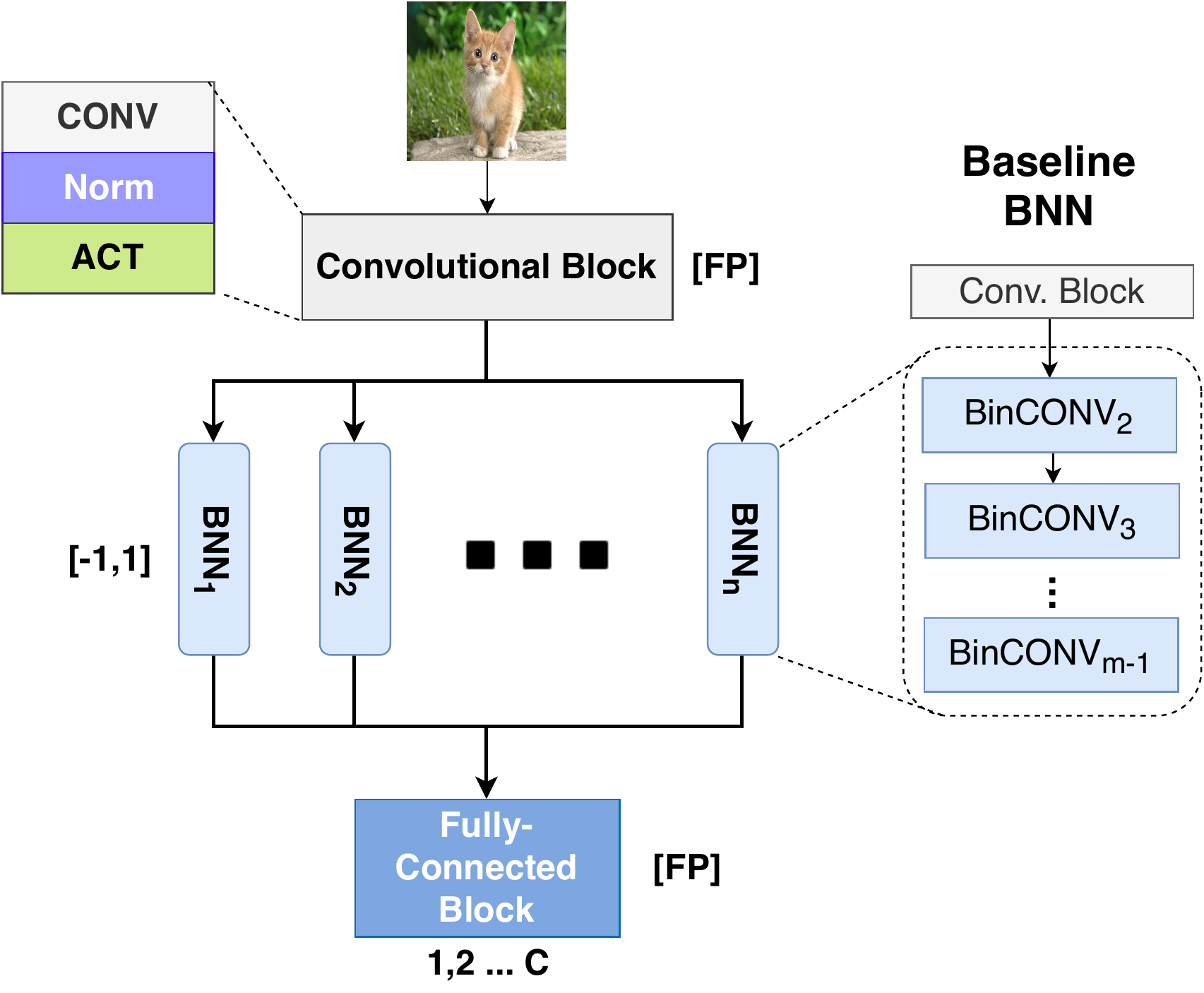}
    \caption{TentacleNet architecture. CONV refers to convolutional layers, Norm to normalization layer, ACT to activation layers. The prefix Bin stands for binary.
    }
    \label{fig:tentaclenet}
\end{figure}

A high-level view is depicted in Fig. \ref{fig:tentaclenet}. The inner core consists of $n$ parallel branches, the {\em tentacles}, which play as the weak estimators. Each tentacle (labeled as BNN$_i$) is a replica of the binarized floating-point model, except for the first and last layer which are shared among all the tentacles, these are the {\em Convolutional Block} and the {\em Fully-Connected Block} in Fig. \ref{fig:tentaclenet}.
The former (grey box) is in charge of producing a common activation map fed as input to all the tentacles. It contains three sub-layers, convolution (CONV), normalization (Norm) and activation (ACT). The latter (blue box) implements the actual classification of the binary features extracted along the tentacles. It is worth emphasizing that all the tentacles operate full binary operations [-1,1], while the {\em Convolutional Block} and the {\em Fully-Connected Block} are taken to full-precision [FP], i.e. floating-point 32-bit. This design choice is inherited from state-of-the-art BNN models \cite{BinCon, BNN, xnor}, which suggest leaving the first and last layer to full-precision gets higher accuracy. Another important aspect is that the sharing of the two blocks, the most memory demanding due to high arithmetic precision, contributes to save memory space.
%The resulting monolithic structure makes TentacleNet end-to-end trainable.

\begin{figure}[!tt]
%\hspace{-1.2cm}
    \includegraphics[scale=0.35]{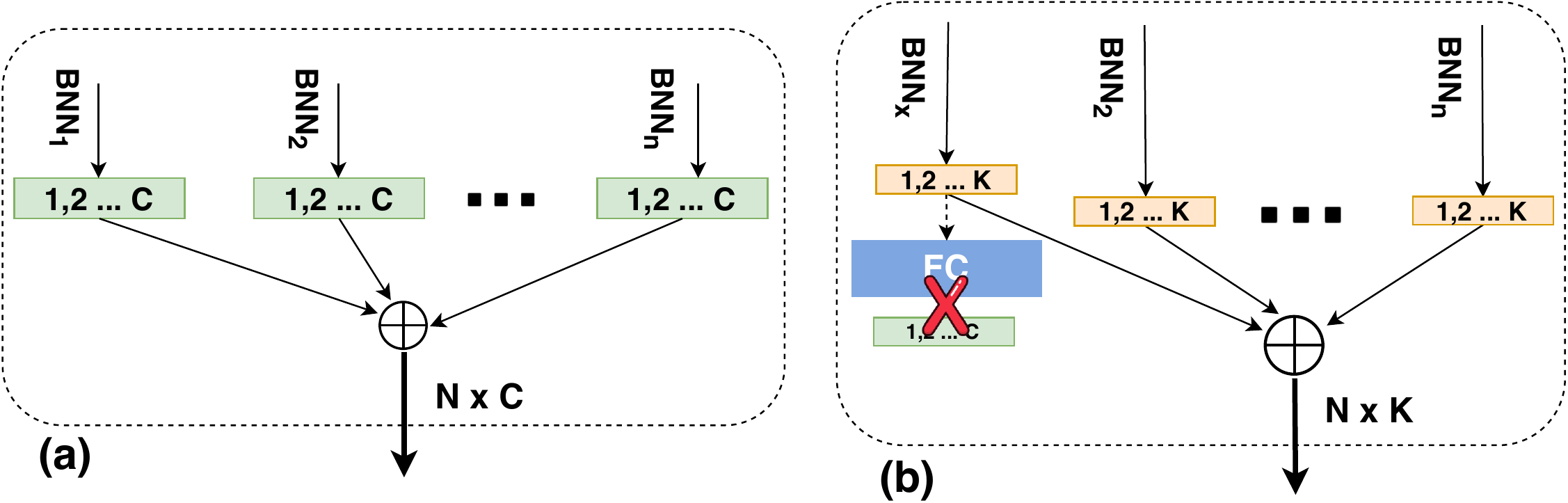}
    \caption{Composition of the Fully-Connected Block}
    \label{fig:fc_scheme}
\end{figure}

The shape of the shared {\em Fully-Connected Block} differs depending on the topology of the original full-precision model. If the original model does produce the $C$ logits through global pooling (where $C$ is the number of classes), namely, without any fully-connected layer (Fig. \ref{fig:fc_scheme}-a), the logits are simply concatenated as a 1-D vector of cardinality $N \times C$ and then fed as input to the {\em Fully-Connected Block}, which is a dense layer of shape $N \times C$ inputs and $ N\times C^2$ weights. Otherwise, if the original model has its own fully-connected layer to produce the logits (Fig. \ref{fig:fc_scheme}-b), we drop it out and concatenate the feature maps as a 1-D vector of cardinality $N \times K$, with $K$ the number of features of each weak estimator; in such case the {\em Fully-Connected Block} is a dense layer of shape $N \times K$ inputs and $ N \times K \times C$ weights.

\subsection{Building and Training Methodology}
TentacleNet can be seen as a pseudo-ensemble that implements some mixed features belonging to the stacking and boosting methods, in particular: the stack is composed of heterogeneous learners with different data-representation; the outputs of the weak estimators are evaluated through a linear transformation; all the layers, including the first and last block, are trained within a single procedure using the same data-set. The result is an end-to-end trainable model whose parameters can be learned through classical back-propagation.
%The resulting monolithic structure makes TentacleNet end-to-end trainable.

%From another perspective, TentacleNet can be thought as a monolithic CNN built upon parallel sub-networks.
The assembling of TentacleNet encompasses few stages. The entry level is a pre-trained floating-point CNN model binarized following the topology described in \cite{xnor}, namely, first and last layers as floating-point and the inner $m-2$ layers as binary. The sequence of such $m-2$ binary layers (from BinCONV$_2$ to BinCONV$_{m-1}$ as reported in the left diagram of Fig.\ref{fig:tentaclenet}) builds a tentacle. $n$ replicas of the same tentacle are placed in parallel (from BNN$_1$ to BNN$_n$ in Fig. \ref{fig:tentaclenet}) and then tied to the top and the bottom with the first convolutional block and the last fully connected block as described in the section \ref{sec:arch}. Once the TentacleNet is assembled, the training procedure described in \cite{xnor} is deployed to learn the weights of the binary tentacles and the weights of the shared layers.

In order to guarantee enough expressive power and reduce the risk of under-/over-fitting, the {\em tentacles} are initialized with different seeds. Saxe et al. \cite{saxe2013} demonstrated that weights initialized with orthonormal or orthogonal bases achieve better performance. In the binary domain, the matrices that satisfy these conditions are called {\em Hadamard matrices}. They are square matrices of order 1, 2, or $4n$, with $n \in \mathbb{N}$; the entries are -1 and 1 and can be generated using Sylvester's method. In order to adapt the Hadamard matrices to the dimensions of the binary kernels within the tentacles, the rows are randomly removed (to reduce the rank) or replicated (to increase the rank). The resulting {\em pseudo}-Hadamard matrices are sub-optimal but still a favorable initialization \cite{BNN}.

\section{Experimental results}
\subsection{Benchmarks and Data-sets}
TentacleNet has been evaluated on the following tasks.\\
\textbf{Image Classification (CIFAR-10/100)} - the standard image classification problem; the data-set contains 60k 32x32 RGB labeled images and can be configured for 10- or 100-class recognition~\cite{cifar}.\\
\textbf{Facial Expression Recognition (FER13)} - emotion recognition from facial expression; the data-set collects 36k 48x48 gray-scale facial images labeled with seven different facial expressions~\cite{fer13}.

Each of the above tasks is implemented through a specialized CNN model as reported in Table~\ref{tab:bench}; the same models work as baseline to build the TentacleNet. The table collects the classification accuracy (\%) and the model size (kB) of the three networks trained in full-precision 32-bit (row FP32) and after binarization (row BNN); here the BNN models refer to XNOR-Net~\cite{xnor}. As expected, BNNs reach remarkable memory reduction (e.g. 24.2$\times$ for ResNet9 on CIFAR-100) at the cost of significant accuracy loss (8.05\% as the worst-case).

%Three CNN models are trained and used as basic block for TentacleNet. The binary models (BNNs) are trained as reported in~\cite{xnor}, starting from the full-precision one.  
%Table~\ref{tab:bench} reports the accuracy of the full-precision and binary-precision model and the corresponding model size. 
%The models in Table~\ref{tab:bench} act as basic block in the TenctacleNet as described in Section~\ref{sec:arch}.

\subsection{Training and Inference Set-Up}
For each task a dedicated TentacleNet is built starting from the  BNN model. The training of TentacleNet iterates for 300 epochs using an adaptive learning rate ($lr$) schedule: $lr$ updated with step 0.1 every 15 consecutive epochs in which the validation loss does not change. Both training and inference stages are implemented using PyTorch (version 1.1.0) and made run on a server powered with 40-core Intel Xeon CPUs and accelerated with the NVIDIA Titan Xp GPU (CUDA v10.0).

%version\footnote{https://github.com/AlexeyAB/darknet} of the popular open-source \texttt{darknet} framework~\cite{darknet}
%Concerning the processing of the binary layers, we adopted a modified version\footnote{https://github.com/AlexeyAB/darknet} of the popular open-source \texttt{darknet} framework~\cite{darknet}.

Since the focus of this paper is on the accuracy-vs-memory tradeoff of binary ensembles, the assessment of hardware-dependent extra-functional metrics, like latency and energy, is left aside as part of future works.

\begin{comment}
The baseline models and TentacleNet are trained with PyTorch (version 1.0), then converted in a format compliant with the popular open-source framework \texttt{darknet}~\cite{darknet}. In particular, we adopted and modified a special version \footnote{https://github.com/AlexeyAB/darknet} that presents a more efficient implementation of BNN.
The evaluation platform is an Intel\textregistered 
Core i7-8700\texttrademark CPU @ 3.20GHz equipped with 12 cores. The memory hierarchy is composed of 16GB DDR4, 383 KB L1, 1.5MB L2, 12MB L3 caches.
\end{comment}

\begin{savenotes}
\begin{table}[!t]
\resizebox{\columnwidth}{!}{
    \centering
    \begin{tabular}{c|c|c|c|c|} \cline{2-5}
    &\multicolumn{1}{|c|}{\multirow{1}{*}{\textbf{Dataset}}}  & CIFAR-10  & CIFAR-100  & FER13 \\  \cline{2-5}
    &\multicolumn{1}{|c|}{\textbf{Baseline Model}}  & NiN \cite{nin} & ResNet9 \cite{resnet}  & FerNet   \\ \hline    
    \multicolumn{1}{|c|}{\multirow{2}{*}{\bf FP32}} &Accuracy (\%) &88.11 &68.25  &65.16\\ 
     \multicolumn{1}{|c|}{} &Model Size (kB)&3778   &19984    &1880 \\ \hline
    \multicolumn{1}{|c|}{\multirow{2}{*}{\textbf{BNN}}} &Accuracy (\%)   &85.20  & 60.20 & 62.86  \\
     \multicolumn{1}{|c|}{}&Model Size (kB)  &181&826  &64 \\ \hline 
    \end{tabular}
    }
    \vspace{3mm}  
    \caption{Benchmarks: Datasets and CNNs\vspace{-3mm}}
    \label{tab:bench}
   
\end{table}
\end{savenotes}

\subsection{Performance assessment}
The objective of this section is twofold: $(i)$  prove that TentacleNet can push binary CNNs towards full-precision accuracy; $(ii)$ show that TentacleNet outperforms existing binary-ensemble methods, both in terms of accuracy and memory. 

\begin{figure}[!t]
    \centering
    \includegraphics[scale=0.5]{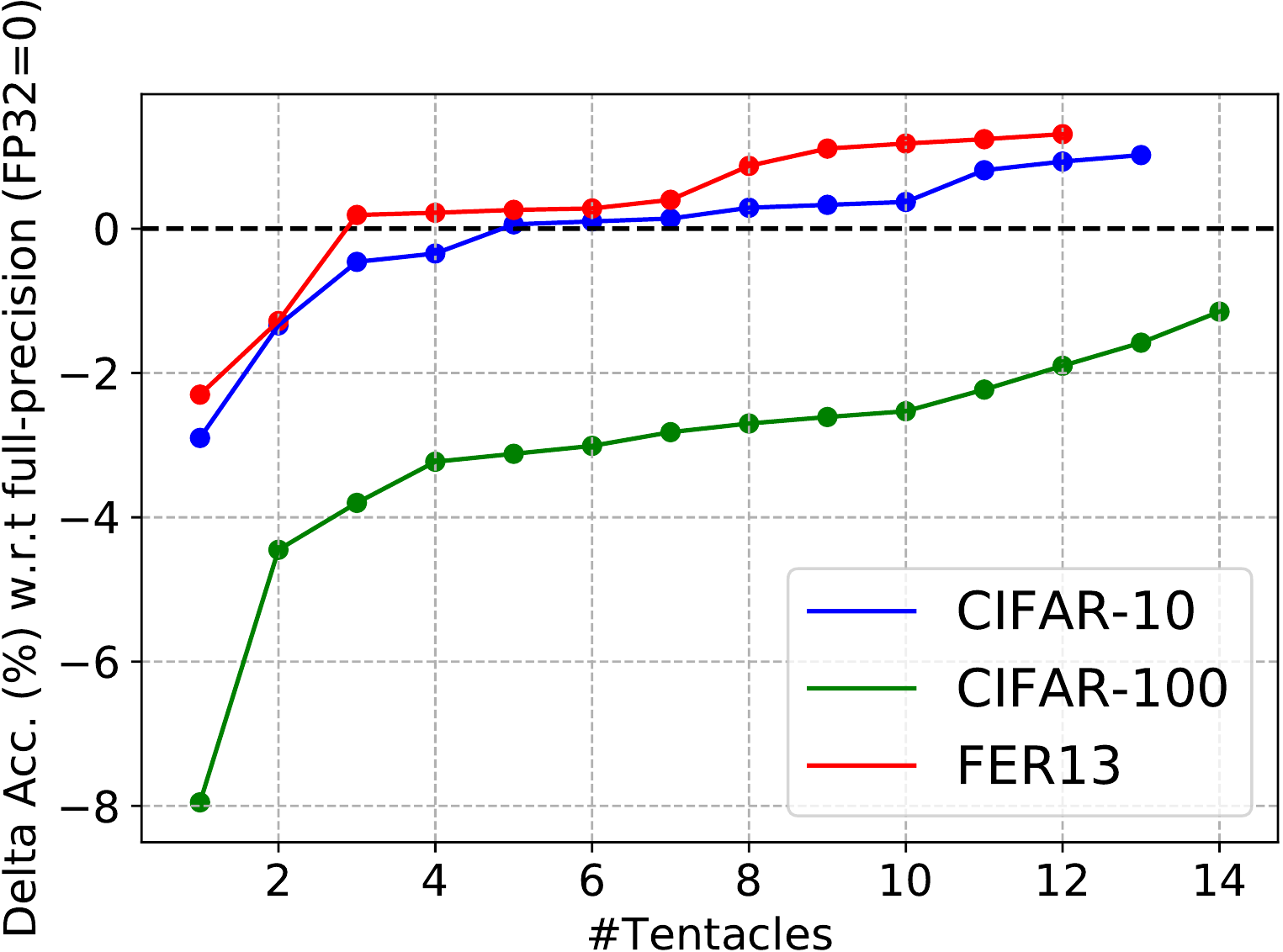}
    \caption{Delta Accuracy (\%) w.r.t. the FP32 model as function of the number of Tentacles.} 
    \label{fig:acc_mem}
\end{figure}

Concerning the first issue, Fig.~\ref{fig:acc_mem} reports a parametric analysis of the classification accuracy achieved by TentacleNet on the three benchmarks. The line plot shows the delta accuracy, which is the difference between TentacleNet and the FP32 model; the break-even point is centered on zero (horizontal dotted line). Deploying just one tentacle, TentacleNet collapses to the original BNN model, namely, the accuracy drop is the same reported in Table~\ref{tab:bench}. As a general trend, the distance to FP32 gets smaller with the number of tentacles. For NiN over CIFAR-10 and FerNet over FER13 TentacleNet reaches the break-even with 3 and 5 tentacles respectively, and it goes even above towards positive values, +1.00\% with 13 tentacles for CIFAR-10 and +1.31\% with 12 tentacles for FER13, meaning that it outperforms FP32 models with much less weight memory: 645kB vs 3778kB for CIFAR-10 (83\% savings), 188kB vs 1880kB for FER13 (90\% savings). The behavior for ResNet over the more complex CIFAR-100 data-set is less performing compared to the other two benchmarks, yet remarkable. With 14 tentacles, the delta accuracy improves from -8.05\% to -1.15\%, very close to FP32, still ensuring low memory footprint, 11465 kB vs 19984 (42.6\% less). For all the three benchmarks, additional experiments revealed the accuracy of TentacleNet saturates, namely, there is no further improvement by increasing the number of tentacles; the top right points of the three lines in the plot of Fig.~\ref{fig:acc_mem} show the highest accuracy that can be reached.
\begin{table}[htt]
 \resizebox{\columnwidth}{!}{
    \centering
   
    \begin{tabular}{|c|cc|c|c|c|}
    
        \hline 
        {\bf Benchmark}&\multicolumn{2}{c|}{\bf Template}&$\mathbf{\Delta}$ {\bf(\%)}&{\bf \shortstack{\#Ensemble/\\ Tentacle}} &{\bf{ \shortstack{M. Size\\ (kB)}}}\\ \hline \hline  \rule{0pt}{2ex}  
        \multirow{3}{*}{\shortstack{CIFAR-10 \\(NiN)}} &\multirow{2}{*}{\cite{binen}}&Bagging &0 &12& 2167 \\ 
        &&Boosting &0 &8&1445  \\ \cline{2-6}
         &&{TentacleNet} &0&5&645 {\bf(55.3\%)}  \\ \hline \hline
         
          \multirow{3}{*}{\shortstack{CIFAR-100 \\(ResNet9)}}  &\multirow{2}{*}{ \cite{binen}} &Bagging &-4.82 &30&24755 \\
          &&   Boosting &-4.77 &25&20629 \\ \cline{2-6}
         &&{TentacleNet} &-1.15 &14&11465 {\bf(44.4\%)} \\ \hline \hline
         
          \multirow{3}{*}{\shortstack{FER13 \\(FerNet)}} & \multirow{2}{*}{ \cite{binen}}  &Bagging &-0.35 &11 &697 \\
           &&   Boosting &-0.67 &26&1648  \\ \cline{2-6}
         &&TentacleNet&0 &3&188 {\bf(73.0\%)} \\\hline
         
    \end{tabular}}
    \vspace{2mm}
    \caption{TentacleNet vs BENN {\cite{binen}}}
    \label{tab:fin_res}
\end{table}
For what concerns the second issue, we provide a quantitative comparison against BENN~\cite{binen}, which is state-of-the-art for binary ensembles. The BENN strategy is to apply standard ensemble methods to BNNs, bagging and boosting in particular. To ensure a fair comparison we implemented and applied the two BENN methods on the benchmarks under analysis. The main results are collected in Table~\ref{tab:fin_res}, which reports the delta accuracy w.r.t. the FP32 model (as in Fig.\ref{fig:acc_mem}), the number of ensembles (for BENN) or tentacles (for TentacleNet), and the memory footprint (the percentage reported in brackets refers to the memory savings of TentacleNet w.r.t. the smallest BENN model). When possible, the comparison is done at the break-even point with the FP32 model (i.e. $\Delta$=0), otherwise at the highest achievable accuracy. TentacleNet is more accurate and more compact than BENN over the three benchmarks. For instance, considering the NiN model over CIFAR-10, both BENN and TentacleNet achieve the accuracy of the FP32 model, but TentacleNet needs less memory to store the weights (55.3\% less w.r.t. boosting). Larger savings have been observed for FerNet over the FER13 benchmark, where BENN and TentacleNet are almost equivalent in terms of accuracy (TentacleNet +0.35\% more accurate than BENN bagging), but with a large memory spread (TentacleNet is 73\% smaller than BENN bagging). Also for the most complex network, namely ResNet over CIFAR-100, TentacleNet reaches higher accuracy (+3.62\% w.r.t. BENN boosting) with large memory savings (44.4\%). To be noted that the memory footprint of the smallest BENN (20629 kB) gets bigger than the original FP32 model (19984 kB). Overall, TentacleNet is more accurate and much smaller than other binary ensemble methods.

\section{Conclusions}

\bibliographystyle{IEEEtran}
\bibliography{refs}

\end{document}